\documentclass[11pt,a4paper]{article}
\usepackage[hyperref]{emnlp-ijcnlp-2019}
\usepackage{times}
\usepackage{latexsym}

\usepackage{url}
\usepackage{graphicx}
\usepackage{float} 
\usepackage{subfigure}
\usepackage{amsmath}
\usepackage{amsfonts}
\usepackage{multirow}
\usepackage{makecell}
\usepackage{enumerate}

\aclfinalcopy 

\title{A Novel Aspect-Guided Deep Transition Model \\ for Aspect Based Sentiment Analysis}
\author{
  Yunlong Liang\textsuperscript{1}\thanks{ \ \ Work was done when Yunlong Liang was an intern at Pattern Recognition Center, WeChat AI, Tencent Inc, China.} ,
  Fandong Meng\textsuperscript{2},
  Jinchao Zhang\textsuperscript{2},
  Jinan Xu\textsuperscript{1}\thanks{ \ \ Jinan Xu is the corresponding author.} ,
  Yufeng Chen\textsuperscript{1}
  and Jie Zhou\textsuperscript{2} \\
  \textsuperscript{1}Beijing Jiaotong University, China \\
  \textsuperscript{2}Pattern Recognition Center, WeChat AI, Tencent Inc, China \\
  \texttt{\{yunlonliang,jaxu,chenyf\}@bjtu.edu.cn} \\
  \texttt{\{fandongmeng,dayerzhang,withtomzhou\}@tencent.com} \\
}

\date{}

\begin{document}
\maketitle
\begin{abstract}
  Aspect based sentiment analysis (ABSA) aims to identify the sentiment polarity towards the given aspect in a sentence, while previous models typically exploit an aspect-independent (weakly associative) encoder for sentence representation generation. In this paper, we propose a novel \textbf{A}spect-\textbf{G}uided \textbf{D}eep \textbf{T}ransition model, named AGDT, which utilizes the given aspect to guide the sentence encoding from scratch with the specially-designed deep transition architecture. Furthermore, an aspect-oriented objective is designed to enforce AGDT to reconstruct the given aspect with the generated sentence representation. In doing so, our AGDT can accurately generate aspect-specific sentence representation, and thus conduct more accurate sentiment predictions. Experimental results on multiple SemEval datasets demonstrate the effectiveness of our proposed approach, which significantly outperforms the best reported results with the same setting\footnote{The code is publicly available at: \url{https://github.com/XL2248/AGDT}}. 
\end{abstract}

\section{Introduction}
\label{sec:introduction}
Aspect based sentiment analysis (ABSA) is a fine-grained task in sentiment analysis, which can provide important sentiment information for other natural language processing (NLP) tasks. There are two different subtasks in ABSA, namely, aspect-category sentiment analysis and aspect-term sentiment analysis
~\citep{Pontiki:14,weixueGCAE:18}. Aspect-category sentiment analysis aims at predicting the sentiment polarity towards the given aspect, which is in predefined several categories and it may not appear in the sentence. For instance, in Table~\ref{tbl:testE}, the aspect-category sentiment analysis is going to predict the sentiment polarity towards the aspect ``{\em food}'', which is not appeared in the sentence. By contrast, the goal of aspect-term sentiment analysis is to predict the sentiment polarity over the aspect term which is a subsequence of the sentence. For instance, the aspect-term sentiment analysis will predict the sentiment polarity towards the aspect term ``{\em The appetizers}'', which is a subsequence of the sentence. Additionally, the number of categories of the aspect term is more than one thousand in the training corpus.

\begin{table}[t!]
\begin{center}
\setlength{\tabcolsep}{1.4mm}{
\begin{tabular}{c|c|c}
\hline 
\textbf{Sentence}  &\multicolumn{2}{l} {\makecell{The appetizers are ok, \\but the service is slow. }}  \\ \hline
\textbf{Aspect-Category}    &  food     &  service \qquad \\ \hline
\textbf{Aspect-Term}    &  The appetizers &  service \qquad  \\ \hline
\textbf{Sentiment Polarity} &  Neutral &  Negative \qquad  \\
\hline
\end{tabular}}
\end{center}
\caption{The instance contains different sentiment polarities towards two aspects.}
\label{tbl:testE}
\end{table}
As shown in Table~\ref{tbl:testE}, sentiment polarity may be different when different aspects are considered. Thus, the given aspect (term) is crucial to ABSA tasks~\citep{jiang-etal-2011-target,Ma:17,WangS:18,DBLP:journals/corr/abs-1905-07719,liang2019context}. Besides, \citet{li2018transformation} show that not all words of a sentence are useful for the sentiment prediction towards a given aspect (term). For instance, when the given aspect is the ``{\em service}'', the words ``{\em appetizers}'' and ``{\em ok}'' are irrelevant for the sentiment prediction.  
Therefore, an aspect-independent (weakly associative) encoder may encode such background words (e.g., ``{\em appetizers}'' and ``{\em ok}'') into the final representation, which may lead to an incorrect prediction. 

Numerous existing models~\cite{Tang:16b,Yi:17,Fan:18,weixueGCAE:18} typically utilize an aspect-independent encoder to generate the sentence representation, and then apply the attention mechanism~\cite{D15-1166} or gating mechanism to conduct feature selection and extraction, while feature selection and extraction may base on noised representations. In addition, some models~\cite{Tang:16a,Wang:16,Majumder:18} simply concatenate the aspect embedding with each word embedding of the sentence, and then leverage conventional Long Short-Term Memories (LSTMs)~\cite{Hochreiter:1997:LSM:1246443.1246450} to generate the sentence representation. However, it is insufficient to exploit the given aspect and conduct potentially complex feature selection and extraction.

To address this issue, we investigate a novel architecture to enhance the capability of feature selection and extraction with the guidance of the given aspect from scratch. Based on the deep transition Gated Recurrent Unit (GRU)~\cite{Cho:14,journals/corr/PascanuGCB13,W17-4710,Meng:19}, an aspect-guided GRU encoder is thus proposed, which utilizes the given aspect to guide the sentence encoding procedure at the very beginning stage. 
In particular, we specially design an aspect-gate for the deep transition GRU to control the information flow of each token input, with the aim of guiding feature selection and extraction from scratch, i.e. sentence representation generation. Furthermore, we design an aspect-oriented objective to enforce our model to reconstruct the given aspect, with the sentence representation generated by the aspect-guided encoder. We name this \textbf{A}spect-\textbf{G}uided \textbf{D}eep \textbf{T}ransition model as AGDT. With all the above contributions, our AGDT can accurately generate an aspect-specific representation for a sentence, and thus conduct more accurate sentiment predictions towards the given aspect.

We evaluate the AGDT on multiple datasets of two subtasks in ABSA. Experimental results demonstrate the effectiveness of our proposed approach. And the AGDT significantly surpasses existing models with the same setting and achieves state-of-the-art performance among the models without using additional features (e.g., BERT~\citep{bert}). Moreover, we also provide empirical and visualization analysis to reveal the advantages of our model. Our contributions can be summarized as follows:
\begin{itemize}
\item We propose an aspect-guided encoder, which utilizes the given aspect to guide the encoding of a sentence from scratch, in order to conduct the aspect-specific feature selection and extraction at the very beginning stage.
\item We propose an aspect-reconstruction approach to further guarantee that the aspect-specific information has been fully embedded into the sentence representation.
\item Our AGDT substantially outperforms previous systems with the same setting, and achieves state-of-the-art results on benchmark datasets compared to those models without leveraging additional features (e.g., BERT).
\end{itemize}

\section{Model Description}
\begin{figure*}[ht]
\centering
  \includegraphics[width = 0.96\textwidth]{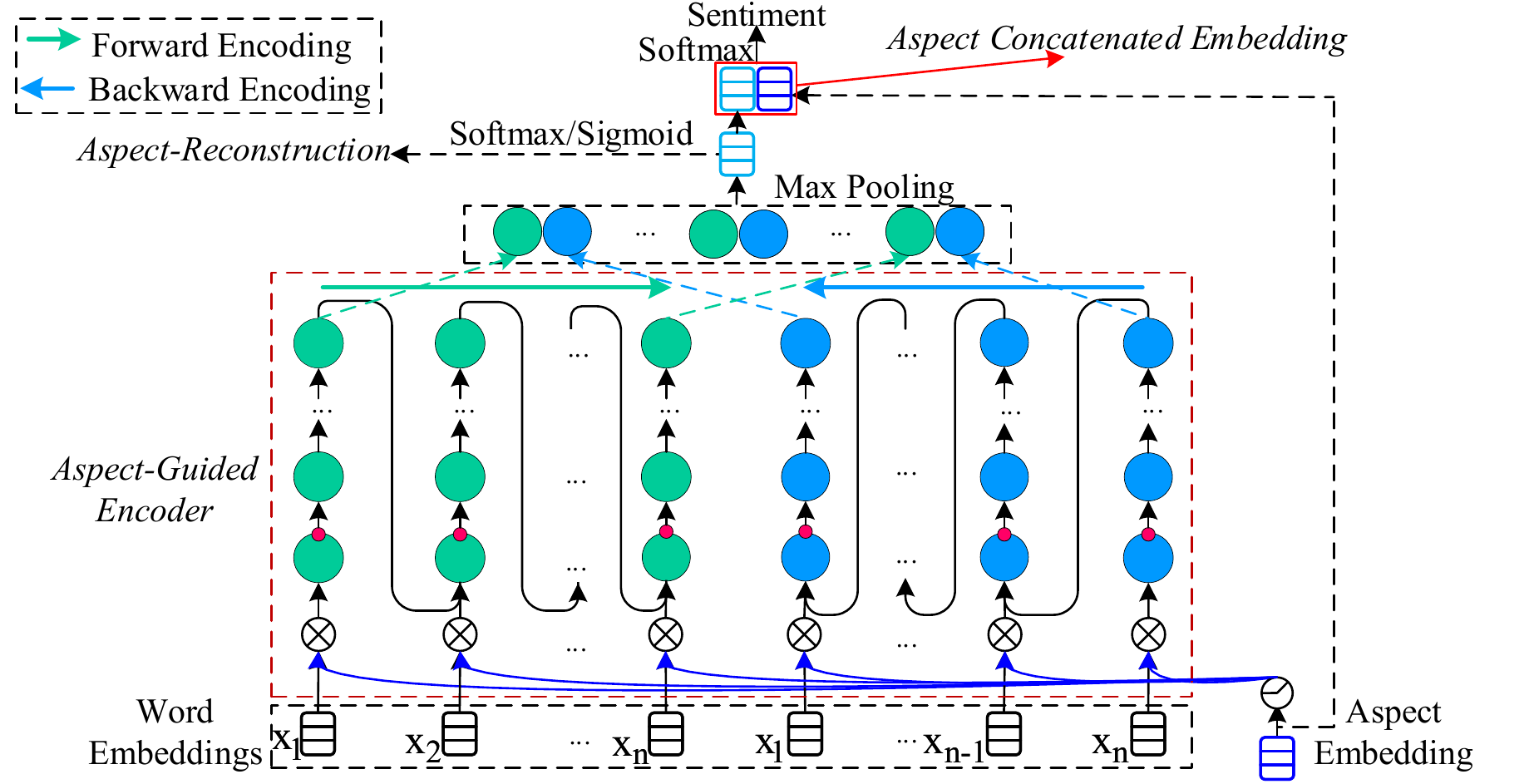}
\caption{The overview of AGDT. The bottom right dark node (above the aspect embedding) is the aspect gate and other dark nodes ($\otimes$) means element-wise multiply for the input token and the aspect gate. The \emph{aspect-guided encoder} consists of a L-GRU (the circle frames fused with a small circle on above) at the bottom followed by several T-GRUs (the circle frames) from bottom to up. }
\label{fig:DTSA_aspect}
\end{figure*}

As shown in Figure~\ref{fig:DTSA_aspect}, the AGDT model mainly consists of three parts: \emph{aspect-guided encoder}, \emph{aspect-reconstruction} and \emph{aspect concatenated embedding}. The aspect-guided encoder is specially designed to guide the encoding of a sentence from scratch for conducting the aspect-specific feature selection and extraction at the very beginning stage. The aspect-reconstruction aims to guarantee that the aspect-specific information has been fully embedded in the sentence representation for more accurate predictions. The aspect concatenated embedding part is used to concatenate the aspect embedding and the generated sentence representation so as to make the final prediction. 

\subsection{Aspect-Guided Encoder}

The aspect-guided encoder is the core module of AGDT, which consists of two key components: Aspect-guided GRU and Transition GRU~\citep{Cho:14}.

\textbf{A-GRU:} Aspect-guided GRU (A-GRU) is a specially-designed unit for the ABSA tasks, which is an extension of the L-GRU proposed by~\citet{Meng:19}. In particular, we design an aspect-gate to select aspect-specific representations through controlling the transformation scale of token embeddings at each time step.

At time step $t$, the hidden state $\mathbf{h}_{t}$ is computed as follows:
\begin{align}
  \label{eq:gru_h_l}
  \mathbf{h}_{t} &= (1 - \mathbf{z}_{t}) \odot \mathbf{h}_{t-1} + \mathbf{z}_{t} \odot \widetilde{\mathbf{h}}_{t}
\end{align}
where  $\odot$ represents element-wise product; $\mathbf{z}_{t}$ is the update gate \citep{Cho:14}; and $\widetilde{\mathbf{h}}_{t}$ is the candidate activation, which is computed as:
\begin{align}
    \nonumber
  \widetilde{\mathbf{h}}_{t} &= \text{tanh}(\mathbf{g}_{t} \odot (\mathbf{W}_{xh}\mathbf{x}_{t})  + \mathbf{r}_{t} \odot (\mathbf{W}_{hh}\mathbf{h}_{t-1})) \\
    \label{eq:l_gru_h_f}
  &\quad+ \mathbf{l}_{t} \odot \textbf{H}_1(\mathbf{x}_{t}) + \mathbf{g}_{t} \odot \textbf{H}_2(\mathbf{x}_{t}) 
\end{align}
where $\mathbf{g}_{t}$ denotes the aspect-gate; $\mathbf{x}_{t}$ represents the input word embedding at time step $t$; $\mathbf{r}_{t}$ is the reset gate \citep{Cho:14}; $\textbf{H}_1(\mathbf{x}_{t})$ and $\textbf{H}_2(\mathbf{x}_{t})$ are the linear transformation of the input $\mathbf{x}_{t}$, and $\mathbf{l}_{t}$ is the linear transformation gate for $\mathbf{x}_{t}$~\cite{Meng:19}. 
$\mathbf{r}_{t}$, $\mathbf{z}_{t}$,  $\mathbf{l}_{t}$, $\mathbf{g}_{t}$, $\textbf{H}_{1}(\mathbf{x}_{t})$ and $\textbf{H}_{2}(\mathbf{x}_{t})$ are computed as: 
\begin{align}
\label{eq:gru_r}
  &\mathbf{r}_{t} = \sigma(\mathbf{W}_{xr}\mathbf{x}_{t} + \mathbf{W}_{hr}\mathbf{h}_{t-1})\\
  \label{eq:gru_z}
  &\mathbf{z}_{t} = \sigma(\mathbf{W}_{xz}\mathbf{x}_{t} + \mathbf{W}_{hz}\mathbf{h}_{t-1})    \\
  \label{eq:gru_l}
  &\mathbf{l}_{t} \, = \sigma(\mathbf{W}_{xl}\mathbf{x}_{t} + \mathbf{W}_{hl}\mathbf{h}_{t-1}) \\
\label{eq:aspect_gate}
  &\mathbf{g}_{t} = \text{relu}(\mathbf{W}_{a}\mathbf{a} + \mathbf{W}_{hg}\mathbf{h}_{t-1}) \\
  \label{eq:H_t1}
  &\mathbf{\textbf{H}_1(\mathbf{x}_{t})} = \mathbf{W}_{1}\mathbf{x}_{t} \\
 \label{eq:H_t2}
  &\mathbf{\textbf{H}_2(\mathbf{x}_{t})} = \mathbf{W}_{2}\mathbf{x}_{t}
\end{align}
where ``$\mathbf{a}$" denotes the embedding of the given aspect, which is the same at each time step. The update gate $\mathbf{z}_t$ and reset gate $\mathbf{r}_t$ are the same as them in the conventional GRU.

In Eq. (\ref{eq:l_gru_h_f}) $\sim$ (\ref{eq:H_t2}), the aspect-gate $\mathbf{g}_{t}$ controls both nonlinear and linear transformations of the input $\mathbf{x}_{t}$ under the guidance of the given aspect at each time step. Besides, we also exploit a linear transformation gate $\mathbf{l}_{t}$ to control the linear transformation of the input, according to the current input $\mathbf{x}_t$ and previous hidden state $\mathbf{h}_{t-1}$, which has been proved powerful in the deep transition architecture~\cite{Meng:19}.

As a consequence, A-GRU can control both non-linear transformation and linear transformation for input $\mathbf{x}_{t}$ at each time step, with the guidance of the given aspect, i.e., A-GRU can guide the encoding of aspect-specific features and block the aspect-irrelevant information at the very beginning stage.

\textbf{T-GRU: } Transition GRU (T-GRU)~\cite{journals/corr/PascanuGCB13} is a crucial component of deep transition block, which is a special case of GRU with only ``state'' as an input, namely its input embedding is zero embedding. As in Figure~\ref{fig:DTSA_aspect}, a deep transition block consists of an A-GRU followed by several T-GRUs at each time step. For the current time step $t$, the output of one A-GRU/T-GRU is fed into the next T-GRU as the input. The output of the last T-GRU at time step $t$ is fed into A-GRU at the time step $t+1$. For a T-GRU, each hidden state at both time step $t$ and transition depth $i$ is computed as: 
\begin{align}
  \label{eq:t_gru_h}
  \mathbf{h}_{t}^i &= (1 - \mathbf{z}_{t}^i) \odot \mathbf{h}_{t}^{i-1} + \mathbf{z}_{t}^i \odot \widetilde{\mathbf{h}}_{t}^i
\\
    \label{eq:t_gru_h_}
  \widetilde{\mathbf{h}}_{t}^i &= \text{tanh}(\mathbf{r}_{t}^i \odot (\mathbf{W}_{h}^i\mathbf{h}_{t}^{i-1}))
\end{align}
where the update gate $\mathbf{z}_{t}^i$ and the reset gate $\mathbf{r}_{t}^i$ are computed as:
\begin{align}
    \label{eq:t_gru_z}
  \mathbf{z}_{t}^i &= \sigma(\mathbf{W}_{z}^i\mathbf{h}_{t}^{i-1}) \\
    \label{eq:t_gru_r}
  \mathbf{r}_{t}^i &= \sigma(\mathbf{W}_{r}^i\mathbf{h}_{t}^{i-1}) 
\end{align}

The AGDT encoder is based on deep transition cells, where each cell is composed of one A-GRU at the bottom, followed by several T-GRUs. Such AGDT model can encode the sentence representation with the guidance of aspect information by utilizing the specially designed architecture. 

\subsection{Aspect-Reconstruction}

We propose an aspect-reconstruction approach to guarantee the aspect-specific information has been fully embedded in the sentence representation. Particularly, we devise two objectives for two subtasks in ABSA respectively. 
In terms of aspect-category sentiment analysis datasets, there are only several predefined aspect categories. While in aspect-term sentiment analysis datasets, the number of categories of term is more than one thousand. In a real-life scenario, the number of term is infinite, while the words that make up terms are limited. Thus we design different loss-functions for these two scenarios.

For the aspect-category sentiment analysis task, we aim to reconstruct the aspect according to the aspect-specific representation. It is a multi-class problem. We take the softmax cross-entropy as the loss function:
\begin{equation}
  \label{eq:aspect_soft_loss}
  \begin{split}
  \mathcal L_{c} &= min (- \sum_{i=0}^{C1}{y}_{i}^{c} \log({p}_{i}^{c}))
  \end{split}
\end{equation}
where C1 is the number of predefined aspects in the training example; ${y}_{i}^{c}$ is the ground-truth and ${p}_{i}^{c}$ is the estimated probability of a aspect.

For the aspect-term sentiment analysis task, we intend to reconstruct the aspect term (may consist of multiple words) according to the aspect-specific representation. It is a multi-label problem and thus the sigmoid cross-entropy is applied:
\begin{equation}
  \begin{split}
  \label{eq:aspect_sigm_loss}
  \mathcal L_{t} &= min\{- \sum_{i=0}^{C2} [{y}_{i}^{t} \log({p}_{i}^{t}) \\
  & \quad + (1-{y}_{i}^{t})\log(1-{p}_{i}^{t})]\}
  \end{split}
\end{equation}
where C2 denotes the number of words that constitute all terms in the training example, ${y}_{i}^{t}$ is the ground-truth and ${p}_{i}^{t}$ represents the predicted value of a word.

Our aspect-oriented objective consists of $\mathcal L_{c}$ and $\mathcal L_{t}$, which guarantee that the aspect-specific information has been fully embedded into the sentence representation.

\subsection{Training Objective}
 
The final loss function is as follows:
\begin{equation}
  \label{eq:J_loss}
  \begin{split}
  \boldsymbol{J}&= min( \underline {- \sum_{i=0}^{C}{y}_{i} \log({p}_{i})} +\boldsymbol{\lambda}  \mathcal L)
  \end{split}
\end{equation}
where the underlined part denotes the conventional loss function; C is the number of sentiment labels; ${y}_{i}$ is the ground-truth and ${p}_{i}$ represents the estimated probability of the sentiment label; $\mathcal L$ is the aspect-oriented objective, where Eq.~\ref{eq:aspect_soft_loss} is for the aspect-category sentiment analysis task and Eq.~\ref{eq:aspect_sigm_loss} is for the aspect-term sentiment analysis task. And $\boldsymbol{\lambda}$ is the weight of $\mathcal L$. 

As shown in Figure~\ref{fig:DTSA_aspect}, we employ the aspect reconstruction approach to reconstruct the aspect (term), where ``softmax'' is for the aspect-category sentiment analysis task and ``sigmoid'' is for the aspect-term sentiment analysis task. Additionally, we concatenate the aspect embedding on the aspect-guided sentence representation to predict the sentiment polarity. Under that loss function (Eq.~\ref{eq:J_loss}), the AGDT can produce aspect-specific sentence representations.

\section{Experiments}
\subsection{Datasets and Metrics}
\label{ssec:ExperSet}
\begin{table*}[t!]
\centering
\setlength{\tabcolsep}{1.50mm}{
\begin{tabular}{l|l|ll|ll|ll|ll|ll}
\hline
\multirow{2}{*}{} & \multicolumn{1}{c|}{} & \multicolumn{2}{c|}{\textbf {Positive}} & \multicolumn{2}{c|}{\textbf {Negative}} & \multicolumn{2}{c|}{\textbf {Neutral}} & \multicolumn{2}{c|}{\textbf{Conflict}} & \multicolumn{2}{c}{\textbf{Total}} \\
\cline{3-12}
                                      &        & DS      & HDS      & DS      & HDS      & DS      & HDS      & DS      & HDS      & DS      & HDS \\ \hline
\multirow{2}{*}{$\textbf{Restaurant-14}$} & Train         & 2,179   & 139        & 839        & 136        & 500        & 50        & 195          & 40    & 3,713   & 365 \\
& Test         & 657      & 32       & 222       & 26        & 94       & 12          & 52       & 19   & 1,025   & 89 \\ \hline
\multirow{2}{*}{$\textbf{Restaurant-Large}$} & Train        & 2,710      & 182       & 1,198       & 178        & 757       & 107          & -   & -    & 4,665   & 467 \\
& Test         & 1,505      & 92       & 680       & 81        & 241       & 61          & -   & -   & 2,426   & 234 \\ \hline

\end{tabular}}
\caption{Statistics of datasets for the aspect-category sentiment analysis task.}
\label{tbl:aspect-category sentiment analysis}
\end{table*}

\begin{table*}[t!]
\centering           
\setlength{\tabcolsep}{1.50mm}{
\begin{tabular}{l|l|ll|ll|ll|ll|ll|ll}
\hline
\multirow{2}{*}{} & \multicolumn{1}{c|}{} & \multicolumn{2}{c|}{\textbf {Positive}} & \multicolumn{2}{c|}{\textbf {Negative}} & \multicolumn{2}{c|}{\textbf {Neutral}} & \multicolumn{2}{c|}{\textbf{Conflict}} & \multicolumn{2}{c|}{\textbf{Total}} & \multicolumn{1}{c}{\textbf{NC}} \\
\cline{3-13}
                 &         & DS      & HDS      & DS      & HDS      & DS      & HDS      & DS      & HDS     & DS      & HDS  & DS\\ \hline
\multirow{2}{*}{$\textbf{Restaurant}$} & Train    & 2,164      & 379       & 805       & 323        & 633       & 293      & 91   & 43    & 3,693   & 1,038 & 3,602 \\ 
& Test    & 728      & 92       & 196       & 62        & 196       & 83          & 14       & 8    & 1,134       & 245 & 1,120\\ \hline
\multirow{2}{*}{$\textbf{Laptop}$} & Train    & 987      & 159       & 866       & 147        & 460       & 173          & 45   & 17     & 2,358   & 496    & 2,313\\ 
& Test     & 341      & 31       & 128       & 25        & 169       & 49          & 16       & 3   & 654       & 108   & 638 \\ \hline
\end{tabular}}
\caption{Statistics of datasets for the aspect-term sentiment analysis task. The `\textbf{NC}' indicates No ``{\em Conflict}'' label, which is just removed the ``{\em conflict}'' label and is prepared for the three-class experiment.}
\label{tbl:aspect-term sentiment analysis}
\end{table*}
\paragraph{Data Preparation.}

We conduct experiments on two datasets of the aspect-category based task and two datasets of the aspect-term based task. For these four datasets, we name the full dataset as ``DS". In each ``DS", there are some sentences like the example in Table~\ref{tbl:testE}, containing different sentiment labels, each of which associates with an aspect (term). For instance, Table~\ref{tbl:testE} shows the customer's different attitude towards two aspects: ``{\em food}'' (``{\em The appetizers}") and ``{\em service}''. In order to measure whether a model can detect different sentiment polarities in one sentence towards different aspects, we extract a hard dataset from each ``DS'', named ``HDS'', in which each sentence only has different sentiment labels associated with different aspects.  
When processing the original sentence \emph{$s$} that has multiple aspects \emph{${a}_{1},{a}_{2},...,{a}_{n}$} and corresponding sentiment labels \emph{${l}_{1},{l}_{2},...,{l}_{n}$} (\emph{$n$} is the number of aspects or terms in a sentence), the sentence will be expanded into (s, ${a}_{1}$, ${l}_{1}$), (s, ${a}_{2}$, ${l}_{2}$), ..., (s, ${a}_{n}$, ${l}_{n}$) in each dataset~\citep{Ruder:16,ruder-etal-2016-hierarchical,weixueGCAE:18}, i.e, there will be \emph{$n$} duplicated sentences associated with different aspects and labels.

\paragraph{Aspect-Category Sentiment Analysis.}
For comparison, we follow~\citet{weixueGCAE:18} and use the restaurant reviews dataset of SemEval 2014 (``restaurant-14'') Task 4~\citep{Pontiki:14} to evaluate our AGDT model. The dataset contains five predefined aspects and four sentiment labels. 
A large dataset (``restaurant-large'') involves restaurant reviews of three years, i.e., 2014 $\sim$ 2016~\citep{Pontiki:14}. There are eight predefined aspects and three labels in that dataset. When creating the ``restaurant-large'' dataset, we follow the same procedure as in~\citet{weixueGCAE:18}. Statistics of datasets are shown in Table~\ref{tbl:aspect-category sentiment analysis}.

\paragraph{Aspect-Term Sentiment Analysis.}
We use the restaurant and laptop review datasets of SemEval 2014 Task 4~\citep{Pontiki:14} to evaluate our model. Both datasets contain four sentiment labels. Meanwhile, we also conduct a three-class experiment, in order to compare with some work~\citep{Wang:16,Ma:17,li2018transformation} which removed ``conflict'' labels.
Statistics of both datasets are shown in Table~\ref{tbl:aspect-term sentiment analysis}.

\paragraph{Metrics.}
The evaluation metrics are accuracy. All instances are shown in Table~\ref{tbl:aspect-category sentiment analysis} and Table~\ref{tbl:aspect-term sentiment analysis}. Each experiment is repeated five times. The mean and the standard deviation are reported.

\subsection{Implementation Details}
We use the pre-trained 300d Glove\footnote{Pre-trained Glove embeddings can be obtained from \url{http://nlp.stanford.edu/projects/glove/}}
embeddings \citep{glove:14} to initialize word embeddings, which is fixed in
all models. For out-of-vocabulary words, we randomly sample their embeddings by the uniform distribution $U(-0.25, 0.25)$. Following \citet{Tang:16b,Chen:17,Liu:17}, we take the averaged word embedding as the aspect representation for multi-word aspect terms. The transition depth of deep transition model is 4 (see Section~\ref{sec:impactofdepth}). The hidden size is set to 300. We set the dropout rate~\citep{dropout:14} to 0.5 for input token embeddings and 0.3 for hidden states. All models are optimized using Adam optimizer~\citep{Adam:14} with gradient clipping equals to 5~\citep{DBLP:journals/corr/abs-1211-5063}. The initial learning rate is set to 0.01 and the batch size is set to 4096 at the token level. The weight of the reconstruction loss $\boldsymbol{\lambda}$ in Eq.~\ref{eq:J_loss} is fine-tuned (see Section~\ref{sec:impactofloss}) and respectively set to 0.4, 0.4, 0.2 and 0.5 for four datasets. 
 
\begin{table*}[t!]
\centering
\begin{tabular}{l|ll|ll}
\hline
\multicolumn{1}{c|}{\multirow{2}{*}{\textbf{Models}}} &\multicolumn{2}{c|}{$\textbf{Restaurant-14}$}  &  \multicolumn{2}{c}{$\textbf{Restaurant-Large}$}                        \\ \cline{2-5}
\multicolumn{1}{c|}{}                    & \multicolumn{1}{c}{DS} & \multicolumn{1}{c|}{HDS} & \multicolumn{1}{c}{DS} & \multicolumn{1}{c}{HDS}        \\ \hline

\textbf{ATAE-LSTM}\citep{Wang:16}*          & 78.29$\pm$0.68     & 45.62$\pm$0.90      & 83.91$\pm$0.49      & 66.32$\pm$2.28                 \\
\textbf{CNN}\citep{DBLP:journals/corr/Kim14f}*                    & 79.47$\pm$0.32     & 44.94$\pm$0.01       & 84.28$\pm$0.15      & 50.43$\pm$0.38            \\

\textbf{GCAE}\citep{weixueGCAE:18}*                & 79.35$\pm$0.34    & 50.55$\pm$1.83   & 85.92$\pm$0.27      & 70.75$\pm$1.19    \\ \hline

\textbf{AGDT}                  & \textbf{81.78}$\pm$0.31    & \textbf{62.02}$\pm$1.31 & \textbf{87.55}$\pm$0.17     & \textbf{75.73}$\pm$0.50  \\ 
\hline
\end{tabular}
\caption{The accuracy of the aspect-category sentiment analysis task. `*' refers to citing from GCAE \citep{weixueGCAE:18}. }
\label{tbl:result_aspect-category sentiment analysis}
\end{table*}

\begin{table*}[t!]
\centering
\begin{tabular}{l|ll|ll}
\hline
\multirow{2}{*}{\textbf{Models}}    & \multicolumn{2}{c|}{$\textbf{Restaurant}$}   & \multicolumn{2}{c}{$\textbf{Laptop}$}             \\ \cline{2-5}
& \multicolumn{1}{c}{DS} & \multicolumn{1}{c|}{HDS} & \multicolumn{1}{c}{DS} & \multicolumn{1}{c}{HDS}     \\ \hline
\textbf{TD-LSTM}\citep{Tang:16a}*               & 73.44$\pm$1.17    & 56.48$\pm$2.46    & 62.23$\pm$0.92    & 46.11$\pm$1.89   \\
\textbf{ATAE-LSTM}\citep{Wang:16}*             & 73.74$\pm$3.01    & 50.98$\pm$2.27    & 64.38$\pm$4.52    & 40.39$\pm$1.30   \\
\textbf{IAN}\citep{Ma:17}*                   & 76.34$\pm$0.27    & 55.16$\pm$1.97    & 68.49$\pm$0.57    & 44.51$\pm$0.48   \\
\textbf{RAM}\citep{Chen:17}*                   & 76.97$\pm$0.64    & 55.85$\pm$1.60    & 68.48$\pm$0.85    & 45.37$\pm$2.03   \\
\textbf{GCAE}\citep{weixueGCAE:18}*                  & 77.28$\pm$0.32    & 56.73$\pm$0.56    & 69.14$\pm$0.32    & 47.06$\pm$2.45        \\ \hline
\textbf{AGDT}                  & \textbf{78.85}$\pm$0.45     & \textbf{60.33}$\pm$1.01      & \textbf{71.50}$\pm$0.85    & \textbf{51.30}$\pm$1.26 \\ 
\hline
\end{tabular}
\caption{The accuracy of the aspect-term sentiment analysis task. `*' refers to citing from GCAE \citep{weixueGCAE:18}. }
\label{tbl:result_aspect-term sentiment analysis}
\end{table*}
\subsection{Baselines}
To comprehensively evaluate our AGDT, we compare the AGDT with several competitive models. 

\textbf{ATAE-LSTM.} It is an attention-based LSTM model. It appends the given aspect embedding with each word embedding, and then the concatenated embedding is taken as the input of LSTM. The output of LSTM is appended aspect embedding again. Furthermore, attention is applied to extract features for final predictions.

\textbf{CNN.} This model focuses on extracting n-gram features to generate sentence representation for the sentiment classification.

\textbf{TD-LSTM.} This model uses two LSTMs to capture the left and right context of the term to generate target-dependent representations for the sentiment prediction.

\textbf{IAN.} This model employs two LSTMs and interactive attention mechanism to learn representations of the sentence and the aspect, and concatenates them for the sentiment prediction.

\textbf{RAM.} This model applies multiple attentions and memory networks to produce the sentence representation. 

\textbf{GCAE.} It uses CNNs to extract features and then employs two Gated Tanh-Relu units to selectively output the sentiment information flow towards the aspect for predicting sentiment labels. 
\subsection{Main Results and Analysis}
\label{ssec:Res}

\subsubsection*{Aspect-Category Sentiment Analysis Task}
We present the overall performance of our model and baseline models in Table~\ref{tbl:result_aspect-category sentiment analysis}. Results show that our AGDT outperforms all baseline models on both ``restaurant-14'' and ``restaurant-large'' datasets. 
ATAE-LSTM employs an aspect-weakly associative encoder to generate the aspect-specific sentence representation by simply concatenating the aspect, which is insufficient to exploit the given aspect. Although GCAE incorporates the gating mechanism to control the sentiment information flow according to the given aspect, the information flow is generated by an aspect-independent encoder. Compared with GCAE, our AGDT improves the performance by 2.4\% and 1.6\% in the ``DS'' part of the two dataset, respectively. These results demonstrate that our AGDT can sufficiently exploit the given aspect to generate the aspect-guided sentence representation, and thus conduct accurate sentiment prediction. Our model benefits from the following aspects. First, our AGDT utilizes an aspect-guided encoder, which leverages the given aspect to guide the sentence encoding from scratch and generates the aspect-guided representation. Second, the AGDT guarantees that the aspect-specific information has been fully embedded in the sentence representation via reconstructing the given aspect. Third, the given aspect embedding is concatenated on the aspect-guided sentence representation for final predictions. 

The ``HDS'', which is designed to measure whether a model can detect different sentiment polarities in a sentence, consists of replicated sentences with different sentiments towards multiple aspects. Our AGDT surpasses GCAE by a very large margin (+\textbf{11.4\%} and +\textbf{4.9\%} respectively) on both datasets. This indicates that the given aspect information is very pivotal to the accurate sentiment prediction, especially when the sentence has different sentiment labels, which is consistent with existing work~\citep{jiang-etal-2011-target,Ma:17,WangS:18}. Those results demonstrate the effectiveness of our model and suggest that our AGDT has better ability to distinguish the different sentiments of multiple aspects compared to GCAE.

\subsubsection*{Aspect-Term Sentiment Analysis Task}
As shown in Table~\ref{tbl:result_aspect-term sentiment analysis}, our AGDT consistently outperforms all compared methods on both domains. In this task, TD-LSTM and ATAE-LSTM use a aspect-weakly associative encoder. IAN, RAM and GCAE employ an aspect-independent encoder. 
In the ``DS'' part, our AGDT model surpasses all baseline models, which shows that the inclusion of A-GRU (aspect-guided encoder), aspect-reconstruction and aspect concatenated embedding has an overall positive impact on the classification process. 

In the ``HDS'' part, the AGDT model obtains +3.6\% higher accuracy than GCAE on the restaurant domain and +4.2\% higher accuracy on the laptop domain, which shows that our AGDT has stronger ability for the multi-sentiment problem against GCAE. These results further demonstrate that our model works well across tasks and datasets.

\subsubsection*{Ablation Study}
We conduct ablation experiments to investigate the impacts of each part in AGDT, where the GRU is stacked with 4 layers. Here ``AC'' represents aspect concatenated embedding 
, ``AG'' stands for A-GRU (Eq. (\ref{eq:gru_h_l}) $\sim$~(\ref{eq:H_t2})) and ``AR'' denotes the aspect-reconstruction (Eq. (\ref{eq:aspect_soft_loss}) $\sim$~(\ref{eq:J_loss})). 

From Table~\ref{tbl:aspect-category sentiment analysis_albated} and Table~\ref{tbl:aspect-term sentiment analysis_albated}, we can conclude:
\begin{table}[t!]
\begin{center}
\setlength{\tabcolsep}{0.6mm}{
\begin{tabular}{l|c|c|c|cc|cc|l}
\hline
\multirow{2}{*}{} & \multirow{2}{*}{\textbf{AC}} & \multirow{2}{*}{\textbf{AG}} & \multirow{2}{*}{\textbf{AR}} & \multicolumn{2}{c|}{$\textbf{Rest-14}$} & \multicolumn{2}{c|}{$\textbf{Rest-Large}$}  &\multirow{2}{*}{} \\
\cline{5-8}
                              &               &          &            & DS             & HDS                  & DS             & HDS  &              \\ \hline
\textbf{GRU}                  & $\surd$       & $\times$ & $\times$    & 80.90          & 53.93 & 86.75          & 68.46                & {\textcircled{\small{1}}}\\ \hline
\multirow{5}{*} {\textbf{DT}} & $\surd$       & $\times$ & $\times$   & 81.74          & 56.63 & 87.54          & 72.39                  & {\textcircled{\small{2}}}\\ 
                              & $\surd$      & $\surd$  & $\times$   & {81.88}          & 60.42 & \textbf{87.72} & 74.81                   & {\textcircled{\small{3}}}\\
                              & $\times$      & $\surd$  & $\times$  & \textbf{81.95}          & 59.33  & {87.68} & 74.44                   & {\textcircled{\small{4}}}\\ 
                              & $\times$      & $\surd$  & $\surd$   & 81.83          & 61.35 & 87.34          & 75.56                  & {\textcircled{\small{5}}}\\ 
                              & $\surd$       & $\surd$  & $\surd$   & 81.78     & \textbf{62.02}  & 87.55          & \textbf{75.73}        & {\textcircled{\small{6}}}\\ \hline
\end{tabular}}
\end{center}
\caption{Ablation study of the AGDT on the aspect-category sentiment analysis task. Here ``AC'', ``AG'' and ``AR'' represent aspect concatenated embedding, A-GRU and aspect-reconstruction, respectively, `$\surd$' and `$\times$' denotes whether to apply the operation. `Rest-14': Restaurant-14,`Rest-Large': Restaurant-Large. }
\label{tbl:aspect-category sentiment analysis_albated}
\end{table}

\begin{table}[t!]
\begin{center}
\setlength{\tabcolsep}{0.6mm}{
\begin{tabular}{l|c|c|c|cc|cc|l}
\hline
\multirow{2}{*}{}  & \multirow{2}{*}{\textbf{AC}} & \multirow{2}{*}{\textbf{AG}} & \multirow{2}{*}{\textbf{AR}} & \multicolumn{2}{c|}{$\textbf{Restaurant}$} & \multicolumn{2}{c|}{$\textbf{Laptop}$}  &\multirow{2}{*}{} \\
\cline{5-8}
                      &              &          &           & DS              & HDS              & DS                & HDS  &                     \\ \hline
\textbf{GRU}                   & $\surd$      & $\times$ & $\times$  & 78.31           & 55.92            & 70.21             & 46.48   & {\textcircled{\small{1}}}\\ \hline
\multirow{5}{*} {\textbf{DT}}  & $\surd$      & $\times$ & $\times$  & 78.36           & 56.24            & 71.07             & 47.59  & {\textcircled{\small{2}}}\\ 
                               & $\surd$     & $\surd$  & $\times$  & 78.77           & 60.14            & 71.42             & 50.83   & {\textcircled{\small{3}}}\\              
                               & $\times$     & $\surd$  & $\times$  & 78.55           & 60.08            & 71.38             & 50.74   & {\textcircled{\small{4}}}\\ 
                                & $\times$     & $\surd$  & $\surd$   & 78.59           & {60.16}         & 71.47             & 51.11   & {\textcircled{\small{5}}}\\ 
                                & $\surd$      & $\surd$  & $\surd$   & \textbf{78.85}  & \textbf{60.33}            & \textbf{71.50}    & \textbf{51.30}  & {\textcircled{\small{6}}}\\ \hline
\end{tabular}}
\end{center}
\caption{Ablation study of the AGDT on the aspect-term sentiment analysis task.} 
\label{tbl:aspect-term sentiment analysis_albated}
\end{table}
\begin{figure}[h]
\centering
  \includegraphics[width = 0.48\textwidth]{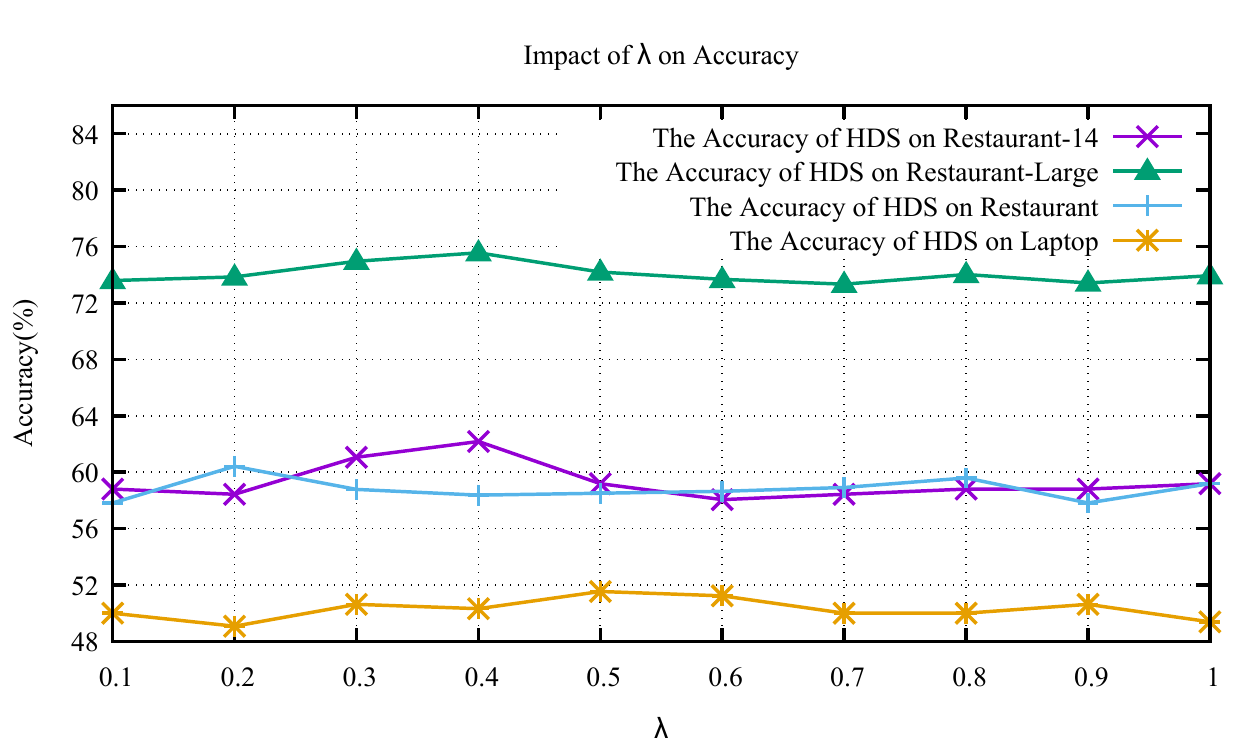}
\caption{The impact of $\boldsymbol{\lambda}$ w.r.t. accuracy on ``HDS''.} 
\label{fig:cee}
\end{figure}
\begin{enumerate}[1).]\setlength{\itemsep}{-0.12cm}
\item 
Deep Transition (DT) achieves superior performances than GRU, which is consistent with previous
work~\citep{W17-4710,Meng:19} 
({\textcircled{\small{2}}} vs. {\textcircled{\small{1}}}).
\item 
Utilizing ``AG'' to guide encoding aspect-related features from scratch has a significant impact for highly competitive results and particularly in the ``HDS'' part, which demonstrates that it has the stronger ability to identify different sentiment polarities towards different aspects. 
({\textcircled{\small{3}}} vs. {\textcircled{\small{2}}}).
\item  Aspect concatenated embedding can promote the accuracy to a degree ({\textcircled{\small{4}}} vs. {\textcircled{\small{3}}}).
\item  
The aspect-reconstruction approach (``AR'') substantially improves the performance, especially in the ``HDS" part ({\textcircled{\small{5}}} vs. {\textcircled{\small{4}}}). 
\item  the results in {\textcircled{\small{6}}} show that all modules have an overall positive impact on the sentiment classification.
\end{enumerate}

\begin{table}[t!]
\centering
\setlength{\tabcolsep}{0.6mm}{
\begin{tabular}{cccccccc}
\hline
& {\textbf{Depth}} &  {\textbf {1}} & {\textbf {2}} & {\textbf {3}} & {\textbf{4}} & {\textbf{5}} & {\textbf{6}}\\
\cline{1-8}

\multirow{2}{*}{{$\mathbb{D}_{\text{1}}$}} & DS & 81.12                & 81.45   & 81.52        & \textbf{81.78}        & 81.07        & 80.68         \\
& HDS         & 55.73   & 57.08        & 60.67        & \textbf{62.02}        & 59.10        & 58.65         \\ \hline
\multirow{2}{*}{{$\mathbb{D}_{\text{2}}$}} & DS        & 87.20   & 87.47        & 87.53        & \textbf{87.55}        & 87.11        & 87.21 \\
& HDS         & 73.93   & 74.27        & \textbf{76.07}        & 75.73        & 75.56        & 74.27  \\ \hline
\multirow{2}{*}{{$\mathbb{D}_{\text{3}}$}} & DS         & 78.18   & 77.94        & 78.69        & \textbf{78.85}        & 78.40        & 77.88      \\
& HDS        & 59.35   & 58.94        & 59.43        & \textbf{60.33}        & 59.27        & 57.80       \\ \hline
\multirow{2}{*}{{$\mathbb{D}_{\text{4}}$}} & DS       & 71.13   & 71.10        & \textbf{71.62}        & 71.50        & 71.16        & 70.86       \\
& HDS         & 49.44   & 50.00        & 50.56        & \textbf{51.30}        & 49.81        & 49.63      \\ \hline

\end{tabular}}
\caption{The accuracy of model depth on the four datasets. `$\mathbb{D}_{\text{1}}$': Restaurant-14, `$\mathbb{D}_{\text{2}}$': Restaurant-Large, `$\mathbb{D}_{\text{3}}$': Restaurant, `$\mathbb{D}_{\text{4}}$': Laptop. }
\label{tbl:model depth}
\end{table}

\begin{table}[t!]
\begin{center}
\setlength{\tabcolsep}{1.5mm}{
\begin{tabular}{l|c|c|c|c}
\hline
& {$\textbf{Rest-14}$} & {$\textbf{Rest-Large}$} & {$\textbf{Rest.}$} &{$\textbf{Laptop}$} \\
\cline{1-5}
DS & 99.55  & 99.80    & 76.21           & 70.92            \\ \hline
\end{tabular}}
\end{center}
\caption{The accuracy of aspect reconstruction on the full test set. `Rest-14': Restaurant-14, `Rest-Large': Restaurant-Large, `Rest.': Restaurant.} 
\label{tbl:accofaspectreconstruction}
\end{table}
\subsubsection*{Impact of Model Depth}
\label{sec:impactofdepth}
 
We have demonstrated the effectiveness of the AGDT. Here, we investigate the impact of model depth of AGDT, varying the depth from 1 to 6. Table~\ref{tbl:model depth} shows the change of accuracy on the test sets as depth increases. We find that the best results can be obtained when the depth is equal to 4 at most case, and further depth do not provide considerable performance improvement.

\subsubsection*{Effectiveness of Aspect-reconstruction Approach}
\label{sec:impactofaspect-reconstruction}
 
Here, we investigate how well the AGDT can reconstruct the aspect information. For the aspect-term reconstruction, we count the construction is correct when all words of the term are reconstructed. Table~\ref{tbl:accofaspectreconstruction} shows all results on four test datasets, which shows the effectiveness of aspect-reconstruction approach again.

\subsubsection*{Impact of Loss Weight $\boldsymbol{\lambda}$}
\label{sec:impactofloss}

We randomly sample a temporary development set from the ``HDS" part of the training set to choose the lambda for each dataset. And we investigate the impact of $\boldsymbol{\lambda}$ for aspect-oriented objectives. Specifically, $\boldsymbol{\lambda}$ is increased from 0.1 to 1.0. Figure~\ref{fig:cee} illustrates all results on four ``HDS" datasets, which show that reconstructing the given aspect can enhance aspect-specific sentiment features and thus obtain better performances.

\subsubsection*{Comparison on Three-Class for the Aspect-Term Sentiment Analysis Task}

We also conduct a three-class experiment to compare our AGDT with previous models, i.e., IARM, TNet, VAE, PBAN, AOA and MGAN, in Table~\ref{tbl:aspect-term sentiment analysis3class}. These previous models are based on an aspect-independent (weakly associative) encoder to generate sentence representations. Results on all domains suggest that our AGDT substantially outperforms most competitive models, except for the TNet on the laptop dataset. The reason may be TNet incorporates additional features (e.g., position features, local ngrams and word-level features) compared to ours (only word-level features).

\begin{table}[t!]
\begin{center}
\setlength{\tabcolsep}{1.2mm}{
\begin{tabular}{l|l|c}
\hline
\multirow{1}{*}{\textbf{Models}} & \multicolumn{1}{c|}{\textbf{$\textbf{Rest.}$}}   & \multicolumn{1}{c}{\textbf{$\textbf{Laptop}$}}             \\ \hline

\textbf{IARM}\citep{Majumder:18}*             & 80.00              & 73.80            \\
\textbf{TNet}\citep{li2018transformation}*             & 80.79              & \textbf{76.54}   \\
\textbf{VAE}\citep{Xu:18}*              & 81.10              & 75.34             \\ 
\textbf{PBAN}\citep{Gu:18}*             & 81.16              & 74.12             \\ 
\textbf{AOA}\citep{Huang:18:AOA:a}*              & 81.20              & 74.50             \\
\textbf{MGAN}\citep{Fan:18}*             & 81.25              & 75.39            \\ 
\textbf{DAuM}\citep{Zhu:18}*             & {82.32}            & 74.45            \\ \hline
\textbf{AGDT}              & \textbf{82.95}     & 75.86            \\ \hline
\end{tabular}}
\end{center}
\caption{The three-class accuracy of the aspect-term sentiment analysis task on SemEval 2014. `*' refers to citing from the original paper. `Rest.': Restaurant.}
\label{tbl:aspect-term sentiment analysis3class}
\end{table}

\section{Analysis and Discussion}        
\label{sec:CSV}

\paragraph{Case Study and Visualization.} 
To give an intuitive understanding of how the proposed A-GRU works from scratch with different aspects, we take a review sentence as an example. 
As the example ``\textit{the appetizers are ok, but the service is slow.}'' shown in Table~\ref{tbl:testE}, it has different sentiment labels towards different aspects. The color depth denotes the semantic relatedness level between the given aspect and each word. More depth means stronger relation to the given aspect.
\begin{figure}[!t]
\centering
  \includegraphics[width = 0.48\textwidth]{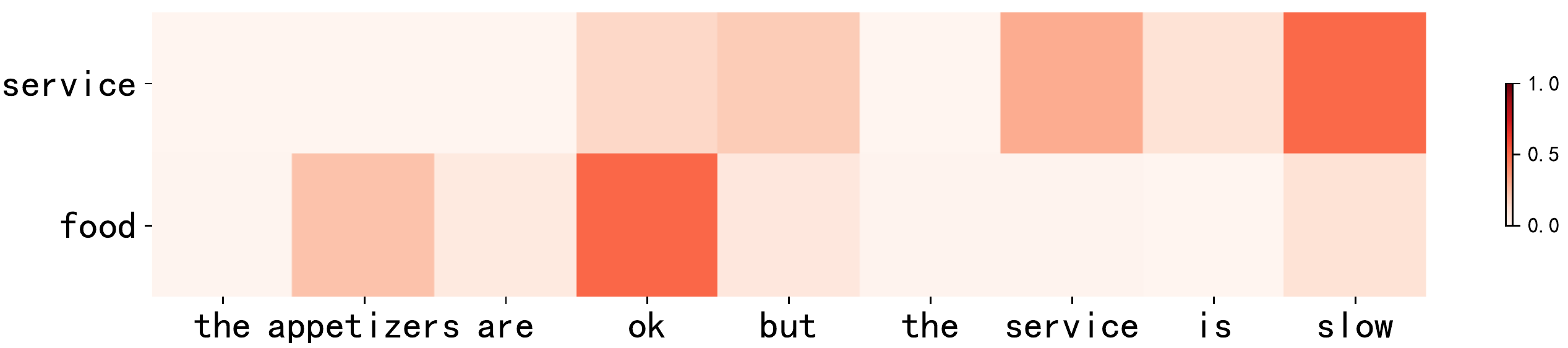}
\caption{The output of A-GRU.}
\label{fig:gate_vis}
\end{figure}
\begin{figure}[!t]
\centering
  \includegraphics[width = 0.48\textwidth]{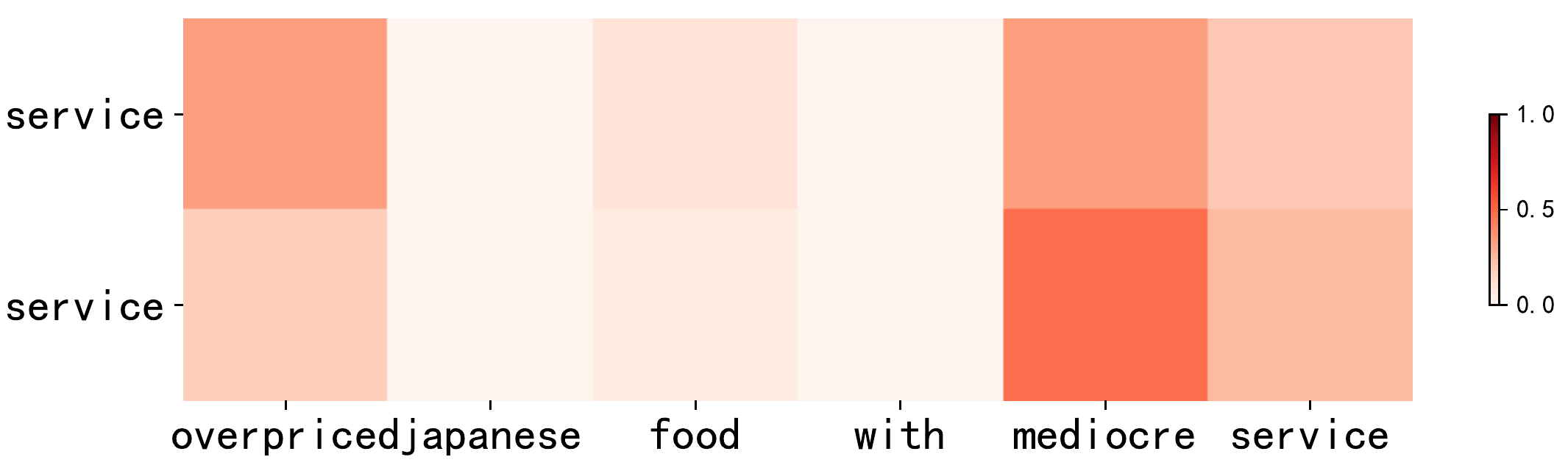}
\caption{The above is the output of A-GRU. The bottom is the output after reconstructing the given aspect. }
\label{fig:gate_w_r}
\end{figure}

Figure~\ref{fig:gate_vis} shows that the A-GRU can effectively guide encoding the aspect-related features with the given aspect and identify corresponding sentiment. In another case, ``\textit{overpriced Japanese food with mediocre service.}'', there are two extremely strong sentiment words. As the above of Figure~\ref{fig:gate_w_r} shows, our A-GRU generates almost the same weight to the word ``{\em overpriced}'' and ``{\em mediocre}''. The bottom of Figure~\ref{fig:gate_w_r} shows that reconstructing the given aspect can effectively enhance aspect-specific sentiment features and produce correct sentiment predictions.

\paragraph{Error Analysis.}
We further investigate the errors from AGDT, which can be roughly divided into 3 types. \textbf{1)} The decision boundary among the sentiment polarity is unclear, even the annotators can not sure what sentiment orientation over the given aspect in the sentence. \textbf{2)} The ``conflict/neutral'' instances are extremely easily misclassified as ``positive'' or ``negative'', due to the imbalanced label distribution in training corpus\footnote{More details can be seen in the dataset or see here: \url{http://alt.qcri.org/semeval2014/}}.
\textbf{3)} The polarity of complex instances is hard to predict, such as the sentence that express subtle emotions, which are hardly effectively captured, or containing negation words (e.g., {\em never}, {\em less} and {\em not}), which easily affect the sentiment polarity. 
\section{Related Work}

\paragraph{Sentiment Analysis.} 
There are kinds of sentiment analysis tasks, such as document-level~\citep{thongtan-phienthrakul-2019-sentiment}, sentence-level\footnote{\url{https://nlp.stanford.edu/sentiment/}}~\citep{zhang-zhang-2019-tree,zhang-etal-2019-latent}, aspect-level~\citep{Pontiki:14,wang-etal-2019-aspect} and multimodal~\citep{8052551,akhtar2019multi} sentiment analysis. For the aspect-level sentiment analysis, previous work typically apply attention mechanism~\citep{D15-1166} combining with memory network~\citep{Jason:WestonCB14} or gating units to solve this task~\citep{Tang:16b,he-etal-2018-effective,Huang:18:PCNN:b,weixueGCAE:18,duan-etal-2018-learning-sentence,Tang:ACL2019,yang2019aspect,bao-etal-2019-attention}, where an aspect-independent encoder is used to generate the sentence representation. In addition, some work leverage the aspect-weakly associative encoder to generate aspect-specific sentence representation~\citep{Tang:16a,Wang:16,Majumder:18}. All of these methods make insufficient use of the given aspect information.
There are also some work which jointly extract the aspect term (and opinion term) and predict its sentiment polarity~\citep{schmitt-etal-2018-joint,DBLP:journals/corr/abs-1811-05082,ma-etal-2018-joint,DBLP:journals/corr/abs-1808-08858,he_acl2019,Luo2019doer,hu2019open,song2019,Wang:2019:ASA:3308558.3313750}. In this paper, we focus on the latter problem and leave
aspect extraction \citep{DBLP:journals/corr/ShuXL17} to future work. And some work~\citep{bert1,bert2,he-etal-2018-exploiting,Xu:18,chen-qian-2019-transfer,he_acl2019} employ the well-known BERT~\citep{bert} or document-level corpora to enhance ABSA tasks, which will be considered in our future work to further improve the performance.

\paragraph{Deep Transition.}

Deep transition has been proved its superiority in language modeling \citep{journals/corr/PascanuGCB13} and machine translation \citep{W17-4710,Meng:19}. We follow the deep transition architecture in \citet{Meng:19} and extend it by incorporating a novel A-GRU for ABSA tasks.
\section{Conclusions }
\label{sec:conclusions}
In this paper, we propose a novel aspect-guided encoder (AGDT) for ABSA tasks, based on a deep transition architecture. Our AGDT can guide the sentence encoding from scratch for the aspect-specific feature selection and extraction. Furthermore, we design an aspect-reconstruction approach to enforce AGDT to reconstruct the given aspect with the generated sentence representation. Empirical studies on four datasets suggest that the AGDT outperforms existing state-of-the-art models substantially on both aspect-category sentiment analysis task and aspect-term sentiment analysis task of ABSA without additional features.

\section*{Acknowledgments}

We sincerely thank the anonymous reviewers for their thorough reviewing and insightful suggestions. Liang, Xu, and Chen are supported by the National Natural Science Foundation of China (Contract 61370130, 61976015, 61473294 and 61876198), and the Beijing Municipal Natural Science Foundation (Contract 4172047), and the International Science and Technology Cooperation Program of the Ministry of Science and Technology (K11F100010).

\bibliography{emnlp-ijcnlp-2019}

\begin{thebibliography}{61}
\expandafter\ifx\csname natexlab\endcsname\relax\def\natexlab#1{#1}\fi

\bibitem[{Akhtar et~al.(2019)Akhtar, Chauhan, Ghosal, Poria, Ekbal, and
  Bhattacharyya}]{akhtar2019multi}
Md~Shad Akhtar, Dushyant Chauhan, Deepanway Ghosal, Soujanya Poria, Asif Ekbal,
  and Pushpak Bhattacharyya. 2019.
\newblock \href {https://doi.org/10.18653/v1/N19-1034} {Multi-task learning for
  multi-modal emotion recognition and sentiment analysis}.
\newblock In \emph{Proceedings of the 2019 Conference of the North {A}merican
  Chapter of the Association for Computational Linguistics: Human Language
  Technologies, Volume 1 (Long and Short Papers)}, pages 370--379, Minneapolis,
  Minnesota. Association for Computational Linguistics.

\bibitem[{Angelidis and Lapata(2018)}]{DBLP:journals/corr/abs-1808-08858}
Stefanos Angelidis and Mirella Lapata. 2018.
\newblock \href {http://arxiv.org/abs/1808.08858} {Summarizing opinions: Aspect
  extraction meets sentiment prediction and they are both weakly supervised}.
\newblock \emph{CoRR}, abs/1808.08858.

\bibitem[{Bao et~al.(2019)Bao, Lambert, and Badia}]{bao-etal-2019-attention}
Lingxian Bao, Patrik Lambert, and Toni Badia. 2019.
\newblock \href {https://www.aclweb.org/anthology/P19-2035} {Attention and
  lexicon regularized {LSTM} for aspect-based sentiment analysis}.
\newblock In \emph{Proceedings of the 57th Annual Meeting of the Association
  for Computational Linguistics: Student Research Workshop}, pages 253--259,
  Florence, Italy. Association for Computational Linguistics.

\bibitem[{{Chen} et~al.(2018){Chen}, {Ji}, {Su}, {Cao}, and {Gao}}]{8052551}
F.~{Chen}, R.~{Ji}, J.~{Su}, D.~{Cao}, and Y.~{Gao}. 2018.
\newblock \href {https://doi.org/10.1109/TMM.2017.2757769} {Predicting
  microblog sentiments via weakly supervised multimodal deep learning}.
\newblock \emph{IEEE Transactions on Multimedia}, 20(4):997--1007.

\bibitem[{Chen et~al.(2017)Chen, Sun, Bing, and Yang}]{Chen:17}
Peng Chen, Zhongqian Sun, Lidong Bing, and Wei Yang. 2017.
\newblock \href {https://doi.org/10.18653/v1/D17-1047} {Recurrent attention
  network on memory for aspect sentiment analysis}.
\newblock In \emph{Proceedings of the 2017 Conference on Empirical Methods in
  Natural Language Processing}, pages 452--461. Association for Computational
  Linguistics.

\bibitem[{Chen and Qian(2019)}]{chen-qian-2019-transfer}
Zhuang Chen and Tieyun Qian. 2019.
\newblock \href {https://www.aclweb.org/anthology/P19-1052} {Transfer capsule
  network for aspect level sentiment classification}.
\newblock In \emph{Proceedings of the 57th Annual Meeting of the Association
  for Computational Linguistics}, pages 547--556, Florence, Italy. Association
  for Computational Linguistics.

\bibitem[{Cho et~al.(2014)Cho, van Merrienboer, Gulcehre, Bahdanau, Bougares,
  Schwenk, and Bengio}]{Cho:14}
Kyunghyun Cho, Bart van Merrienboer, Caglar Gulcehre, Dzmitry Bahdanau, Fethi
  Bougares, Holger Schwenk, and Yoshua Bengio. 2014.
\newblock \href {https://doi.org/10.3115/v1/D14-1179} {Learning phrase
  representations using rnn encoder--decoder for statistical machine
  translation}.
\newblock In \emph{Proceedings of the 2014 Conference on Empirical Methods in
  Natural Language Processing (EMNLP)}, pages 1724--1734. Association for
  Computational Linguistics.

\bibitem[{Dai and Song(2019)}]{song2019}
Hongliang Dai and Yangqiu Song. 2019.
\newblock \href {https://www.aclweb.org/anthology/P19-1520} {Neural aspect and
  opinion term extraction with mined rules as weak supervision}.
\newblock In \emph{Proceedings of the 57th Annual Meeting of the Association
  for Computational Linguistics}, pages 5268--5277, Florence, Italy.
  Association for Computational Linguistics.

\bibitem[{Devlin et~al.(2018)Devlin, Chang, Lee, and Toutanova}]{bert}
Jacob Devlin, Ming{-}Wei Chang, Kenton Lee, and Kristina Toutanova. 2018.
\newblock \href {http://arxiv.org/abs/1810.04805} {{BERT:} pre-training of deep
  bidirectional transformers for language understanding}.
\newblock \emph{CoRR}, abs/1810.04805.

\bibitem[{Duan et~al.(2018)Duan, Ding, and
  Liu}]{duan-etal-2018-learning-sentence}
Junwen Duan, Xiao Ding, and Ting Liu. 2018.
\newblock \href {https://doi.org/10.18653/v1/N18-1051} {Learning sentence
  representations over tree structures for target-dependent classification}.
\newblock In \emph{Proceedings of the 2018 Conference of the North {A}merican
  Chapter of the Association for Computational Linguistics: Human Language
  Technologies, Volume 1 (Long Papers)}, pages 551--560, New Orleans,
  Louisiana. Association for Computational Linguistics.

\bibitem[{Fan et~al.(2018)Fan, Feng, and Zhao}]{Fan:18}
Feifan Fan, Yansong Feng, and Dongyan Zhao. 2018.
\newblock \href {http://aclweb.org/anthology/D18-1380} {Multi-grained attention
  network for aspect-level sentiment classification}.
\newblock In \emph{Proceedings of the 2018 Conference on Empirical Methods in
  Natural Language Processing}, pages 3433--3442. Association for Computational
  Linguistics.

\bibitem[{Gu et~al.(2018)Gu, Zhang, Hou, and Song}]{Gu:18}
Shuqin Gu, Lipeng Zhang, Yuexian Hou, and Yin Song. 2018.
\newblock \href {http://aclweb.org/anthology/C18-1066} {A position-aware
  bidirectional attention network for aspect-level sentiment analysis}.
\newblock In \emph{Proceedings of the 27th International Conference on
  Computational Linguistics}, pages 774--784. Association for Computational
  Linguistics.

\bibitem[{He et~al.(2018{\natexlab{a}})He, Lee, Ng, and
  Dahlmeier}]{he-etal-2018-effective}
Ruidan He, Wee~Sun Lee, Hwee~Tou Ng, and Daniel Dahlmeier. 2018{\natexlab{a}}.
\newblock \href {https://www.aclweb.org/anthology/C18-1096} {Effective
  attention modeling for aspect-level sentiment classification}.
\newblock In \emph{Proceedings of the 27th International Conference on
  Computational Linguistics}, pages 1121--1131, Santa Fe, New Mexico, USA.
  Association for Computational Linguistics.

\bibitem[{He et~al.(2018{\natexlab{b}})He, Lee, Ng, and
  Dahlmeier}]{he-etal-2018-exploiting}
Ruidan He, Wee~Sun Lee, Hwee~Tou Ng, and Daniel Dahlmeier. 2018{\natexlab{b}}.
\newblock \href {https://doi.org/10.18653/v1/P18-2092} {Exploiting document
  knowledge for aspect-level sentiment classification}.
\newblock In \emph{Proceedings of the 56th Annual Meeting of the Association
  for Computational Linguistics (Volume 2: Short Papers)}, pages 579--585,
  Melbourne, Australia. Association for Computational Linguistics.

\bibitem[{He et~al.(2019)He, Lee, Ng, and Dahlmeier}]{he_acl2019}
Ruidan He, Wee~Sun Lee, Hwee~Tou Ng, and Daniel Dahlmeier. 2019.
\newblock \href {https://www.aclweb.org/anthology/P19-1048} {An interactive
  multi-task learning network for end-to-end aspect-based sentiment analysis}.
\newblock In \emph{Proceedings of the 57th Annual Meeting of the Association
  for Computational Linguistics}, pages 504--515, Florence, Italy. Association
  for Computational Linguistics.

\bibitem[{Hochreiter and
  Schmidhuber(1997)}]{Hochreiter:1997:LSM:1246443.1246450}
Sepp Hochreiter and J\"{u}rgen Schmidhuber. 1997.
\newblock \href {https://doi.org/10.1162/neco.1997.9.8.1735} {Long short-term
  memory}.
\newblock \emph{Neural Comput.}, 9(8):1735--1780.

\bibitem[{Hu et~al.(2019)Hu, Peng, Huang, Li, and Lv}]{hu2019open}
Minghao Hu, Yuxing Peng, Zhen Huang, Dongsheng Li, and Yiwei Lv. 2019.
\newblock \href {https://www.aclweb.org/anthology/P19-1051} {Open-domain
  targeted sentiment analysis via span-based extraction and classification}.
\newblock In \emph{Proceedings of the 57th Annual Meeting of the Association
  for Computational Linguistics}, pages 537--546, Florence, Italy. Association
  for Computational Linguistics.

\bibitem[{Huang and Carley(2018)}]{Huang:18:PCNN:b}
Binxuan Huang and Kathleen Carley. 2018.
\newblock \href {http://aclweb.org/anthology/D18-1136} {Parameterized
  convolutional neural networks for aspect level sentiment classification}.
\newblock In \emph{Proceedings of the 2018 Conference on Empirical Methods in
  Natural Language Processing}, pages 1091--1096. Association for Computational
  Linguistics.

\bibitem[{Huang et~al.(2018)Huang, Ou, and Carley}]{Huang:18:AOA:a}
Binxuan Huang, Yanglan Ou, and Kathleen~M. Carley. 2018.
\newblock \href {http://arxiv.org/abs/1804.06536} {Aspect level sentiment
  classification with attention-over-attention neural networks}.
\newblock \emph{CoRR}, abs/1804.06536.

\bibitem[{Jiang et~al.(2011)Jiang, Yu, Zhou, Liu, and
  Zhao}]{jiang-etal-2011-target}
Long Jiang, Mo~Yu, Ming Zhou, Xiaohua Liu, and Tiejun Zhao. 2011.
\newblock \href {https://www.aclweb.org/anthology/P11-1016} {Target-dependent
  twitter sentiment classification}.
\newblock In \emph{Proceedings of the 49th Annual Meeting of the Association
  for Computational Linguistics: Human Language Technologies}, pages 151--160,
  Portland, Oregon, USA. Association for Computational Linguistics.

\bibitem[{Kim(2014)}]{DBLP:journals/corr/Kim14f}
Yoon Kim. 2014.
\newblock \href {http://arxiv.org/abs/1408.5882} {Convolutional neural networks
  for sentence classification}.
\newblock \emph{CoRR}, abs/1408.5882.

\bibitem[{Kingma and Ba(2014)}]{Adam:14}
Diederik~P. Kingma and Jimmy Ba. 2014.
\newblock \href {http://arxiv.org/abs/1412.6980} {Adam: {A} method for
  stochastic optimization}.
\newblock \emph{CoRR}, abs/1412.6980.

\bibitem[{Li et~al.(2018{\natexlab{a}})Li, Bing, Lam, and
  Shi}]{li2018transformation}
Xin Li, Lidong Bing, Wai Lam, and Bei Shi. 2018{\natexlab{a}}.
\newblock \href {http://aclweb.org/anthology/P18-1087} {Transformation networks
  for target-oriented sentiment classification}.
\newblock In \emph{Proceedings of the 56th Annual Meeting of the Association
  for Computational Linguistics (Volume 1: Long Papers)}, pages 946--956.
  Association for Computational Linguistics.

\bibitem[{Li et~al.(2018{\natexlab{b}})Li, Bing, Li, and
  Lam}]{DBLP:journals/corr/abs-1811-05082}
Xin Li, Lidong Bing, Piji Li, and Wai Lam. 2018{\natexlab{b}}.
\newblock \href {http://arxiv.org/abs/1811.05082} {A unified model for opinion
  target extraction and target sentiment prediction}.
\newblock \emph{CoRR}, abs/1811.05082.

\bibitem[{Liang et~al.(2019)Liang, Du, Xu, Li, and Huang}]{liang2019context}
Bin Liang, Jiachen Du, Ruifeng Xu, Binyang Li, and Hejiao Huang. 2019.
\newblock \href {https://www.aclweb.org/anthology/P19-1462} {Context-aware
  embedding for targeted aspect-based sentiment analysis}.
\newblock In \emph{Proceedings of the 57th Annual Meeting of the Association
  for Computational Linguistics}, pages 4678--4683, Florence, Italy.
  Association for Computational Linguistics.

\bibitem[{Liu and Zhang(2017)}]{Liu:17}
Jiangming Liu and Yue Zhang. 2017.
\newblock \href {http://aclweb.org/anthology/E17-2091} {Attention modeling for
  targeted sentiment}.
\newblock In \emph{Proceedings of the 15th Conference of the European Chapter
  of the Association for Computational Linguistics: Volume 2, Short Papers},
  pages 572--577. Association for Computational Linguistics.

\bibitem[{Luo et~al.(2019)Luo, Li, Liu, and Zhang}]{Luo2019doer}
Huaishao Luo, Tianrui Li, Bing Liu, and Junbo Zhang. 2019.
\newblock \href {https://www.aclweb.org/anthology/P19-1056} {{DOER}: Dual
  cross-shared {RNN} for aspect term-polarity co-extraction}.
\newblock In \emph{Proceedings of the 57th Annual Meeting of the Association
  for Computational Linguistics}, pages 591--601, Florence, Italy. Association
  for Computational Linguistics.

\bibitem[{Luong et~al.(2015)Luong, Pham, and Manning}]{D15-1166}
Thang Luong, Hieu Pham, and Christopher~D. Manning. 2015.
\newblock \href {https://doi.org/10.18653/v1/D15-1166} {Effective approaches to
  attention-based neural machine translation}.
\newblock In \emph{Proceedings of the 2015 Conference on Empirical Methods in
  Natural Language Processing}, pages 1412--1421. Association for Computational
  Linguistics.

\bibitem[{Ma et~al.(2018)Ma, Li, and Wang}]{ma-etal-2018-joint}
Dehong Ma, Sujian Li, and Houfeng Wang. 2018.
\newblock \href {https://www.aclweb.org/anthology/D18-1504} {Joint learning for
  targeted sentiment analysis}.
\newblock In \emph{Proceedings of the 2018 Conference on Empirical Methods in
  Natural Language Processing}, pages 4737--4742, Brussels, Belgium.
  Association for Computational Linguistics.

\bibitem[{Ma et~al.(2017)Ma, Li, Zhang, and Wang}]{Ma:17}
Dehong Ma, Sujian Li, Xiaodong Zhang, and Houfeng Wang. 2017.
\newblock \href {http://dl.acm.org/citation.cfm?id=3171837.3171854}
  {Interactive attention networks for aspect-level sentiment classification}.
\newblock In \emph{Proceedings of the 26th International Joint Conference on
  Artificial Intelligence}, IJCAI'17, pages 4068--4074. AAAI Press.

\bibitem[{Majumder et~al.(2018)Majumder, Poria, Gelbukh, Akhtar, Cambria, and
  Ekbal}]{Majumder:18}
Navonil Majumder, Soujanya Poria, Alexander Gelbukh, Md~Shad Akhtar, Erik
  Cambria, and Asif Ekbal. 2018.
\newblock \href {http://aclweb.org/anthology/D18-1377} {Iarm: Inter-aspect
  relation modeling with memory networks in aspect-based sentiment analysis}.
\newblock In \emph{Proceedings of the 2018 Conference on Empirical Methods in
  Natural Language Processing}, pages 3402--3411. Association for Computational
  Linguistics.

\bibitem[{Meng and Zhang(2019)}]{Meng:19}
Fandong Meng and Jinchao Zhang. 2019.
\newblock \href {http://arxiv.org/abs/1812.07807} {{DTMT:} {A} novel deep
  transition architecture for neural machine translation}.
\newblock \emph{CoRR}, abs/1812.07807.

\bibitem[{Miceli~Barone et~al.(2017)Miceli~Barone, Helcl, Sennrich, Haddow, and
  Birch}]{W17-4710}
Antonio~Valerio Miceli~Barone, Jind{\v{r}}ich Helcl, Rico Sennrich, Barry
  Haddow, and Alexandra Birch. 2017.
\newblock \href {https://doi.org/10.18653/v1/W17-4710} {Deep architectures for
  neural machine translation}.
\newblock In \emph{Proceedings of the Second Conference on Machine
  Translation}, pages 99--107. Association for Computational Linguistics.

\bibitem[{Pascanu et~al.(2014)Pascanu, Gülçehre, Cho, and
  Bengio}]{journals/corr/PascanuGCB13}
Razvan Pascanu, Çaglar Gülçehre, Kyunghyun Cho, and Yoshua Bengio. 2014.
\newblock \href
  {http://dblp.uni-trier.de/db/journals/corr/corr1312.html#PascanuGCB13} {How
  to construct deep recurrent neural networks.}
\newblock \emph{CoRR}, abs/1312.6026.

\bibitem[{Pascanu et~al.(2012)Pascanu, Mikolov, and
  Bengio}]{DBLP:journals/corr/abs-1211-5063}
Razvan Pascanu, Tomas Mikolov, and Yoshua Bengio. 2012.
\newblock \href {http://arxiv.org/abs/1211.5063} {Understanding the exploding
  gradient problem}.
\newblock \emph{CoRR}, abs/1211.5063.

\bibitem[{Pennington et~al.(2014)Pennington, Socher, and Manning}]{glove:14}
Jeffrey Pennington, Richard Socher, and Christopher Manning. 2014.
\newblock \href {https://doi.org/10.3115/v1/D14-1162} {Glove: Global vectors
  for word representation}.
\newblock In \emph{Proceedings of the 2014 Conference on Empirical Methods in
  Natural Language Processing (EMNLP)}, pages 1532--1543. Association for
  Computational Linguistics.

\bibitem[{Pontiki et~al.(2014)Pontiki, Galanis, Pavlopoulos, Papageorgiou,
  Androutsopoulos, and Manandhar}]{Pontiki:14}
Maria Pontiki, Dimitris Galanis, John Pavlopoulos, Harris Papageorgiou, Ion
  Androutsopoulos, and Suresh Manandhar. 2014.
\newblock \href {https://doi.org/10.3115/v1/S14-2004} {Semeval-2014 task 4:
  Aspect based sentiment analysis}.
\newblock In \emph{Proceedings of the 8th International Workshop on Semantic
  Evaluation (SemEval 2014)}, pages 27--35. Association for Computational
  Linguistics.

\bibitem[{Ruder et~al.(2016{\natexlab{a}})Ruder, Ghaffari, and
  Breslin}]{ruder-etal-2016-hierarchical}
Sebastian Ruder, Parsa Ghaffari, and John~G. Breslin. 2016{\natexlab{a}}.
\newblock \href {https://doi.org/10.18653/v1/D16-1103} {A hierarchical model of
  reviews for aspect-based sentiment analysis}.
\newblock In \emph{Proceedings of the 2016 Conference on Empirical Methods in
  Natural Language Processing}, pages 999--1005, Austin, Texas. Association for
  Computational Linguistics.

\bibitem[{Ruder et~al.(2016{\natexlab{b}})Ruder, Ghaffari, and
  Breslin}]{Ruder:16}
Sebastian Ruder, Parsa Ghaffari, and John~G. Breslin. 2016{\natexlab{b}}.
\newblock \href {https://doi.org/10.18653/v1/S16-1053} {Insight-1 at
  semeval-2016 task 5: Deep learning for multilingual aspect-based sentiment
  analysis}.
\newblock In \emph{Proceedings of the 10th International Workshop on Semantic
  Evaluation (SemEval-2016)}, pages 330--336. Association for Computational
  Linguistics.

\bibitem[{Schmitt et~al.(2018)Schmitt, Steinheber, Schreiber, and
  Roth}]{schmitt-etal-2018-joint}
Martin Schmitt, Simon Steinheber, Konrad Schreiber, and Benjamin Roth. 2018.
\newblock \href {https://www.aclweb.org/anthology/D18-1139} {Joint aspect and
  polarity classification for aspect-based sentiment analysis with end-to-end
  neural networks}.
\newblock In \emph{Proceedings of the 2018 Conference on Empirical Methods in
  Natural Language Processing}, pages 1109--1114, Brussels, Belgium.
  Association for Computational Linguistics.

\bibitem[{Shu et~al.(2017)Shu, Xu, and Liu}]{DBLP:journals/corr/ShuXL17}
Lei Shu, Hu~Xu, and Bing Liu. 2017.
\newblock \href {http://arxiv.org/abs/1705.00251} {Lifelong learning {CRF} for
  supervised aspect extraction}.
\newblock \emph{CoRR}, abs/1705.00251.

\bibitem[{Srivastava et~al.(2014)Srivastava, Hinton, Krizhevsky, Sutskever, and
  Salakhutdinov}]{dropout:14}
Nitish Srivastava, Geoffrey Hinton, Alex Krizhevsky, Ilya Sutskever, and Ruslan
  Salakhutdinov. 2014.
\newblock \href {http://dl.acm.org/citation.cfm?id=2627435.2670313} {Dropout: A
  simple way to prevent neural networks from overfitting}.
\newblock \emph{J. Mach. Learn. Res.}, 15(1):1929--1958.

\bibitem[{Sun et~al.(2019)Sun, Huang, and Qiu}]{bert1}
Chi Sun, Luyao Huang, and Xipeng Qiu. 2019.
\newblock \href {http://arxiv.org/abs/1903.09588} {Utilizing {BERT} for
  aspect-based sentiment analysis via constructing auxiliary sentence}.
\newblock \emph{CoRR}, abs/1903.09588.

\bibitem[{Tang et~al.(2016{\natexlab{a}})Tang, Qin, Feng, and Liu}]{Tang:16a}
Duyu Tang, Bing Qin, Xiaocheng Feng, and Ting Liu. 2016{\natexlab{a}}.
\newblock \href {http://aclweb.org/anthology/C16-1311} {Effective lstms for
  target-dependent sentiment classification}.
\newblock In \emph{Proceedings of COLING 2016, the 26th International
  Conference on Computational Linguistics: Technical Papers}, pages 3298--3307.
  The COLING 2016 Organizing Committee.

\bibitem[{Tang et~al.(2016{\natexlab{b}})Tang, Qin, and Liu}]{Tang:16b}
Duyu Tang, Bing Qin, and Ting Liu. 2016{\natexlab{b}}.
\newblock \href {https://doi.org/10.18653/v1/D16-1021} {Aspect level sentiment
  classification with deep memory network}.
\newblock In \emph{Proceedings of the 2016 Conference on Empirical Methods in
  Natural Language Processing}, pages 214--224. Association for Computational
  Linguistics.

\bibitem[{Tang et~al.(2019)Tang, Lu, Su, Ge, Song, Sun, and Luo}]{Tang:ACL2019}
Jialong Tang, Ziyao Lu, Jinsong Su, Yubin Ge, Linfeng Song, Le~Sun, and Jiebo
  Luo. 2019.
\newblock \href {https://www.aclweb.org/anthology/P19-1053} {Progressive
  self-supervised attention learning for aspect-level sentiment analysis}.
\newblock In \emph{Proceedings of the 57th Annual Meeting of the Association
  for Computational Linguistics}, pages 557--566, Florence, Italy. Association
  for Computational Linguistics.

\bibitem[{Tay et~al.(2017)Tay, Luu, and Hui}]{Yi:17}
Yi~Tay, Anh~Tuan Luu, and Siu~Cheung Hui. 2017.
\newblock \href {http://arxiv.org/abs/1712.05403} {Learning to attend via
  word-aspect associative fusion for aspect-based sentiment analysis}.
\newblock \emph{CoRR}, abs/1712.05403.

\bibitem[{Thongtan and
  Phienthrakul(2019)}]{thongtan-phienthrakul-2019-sentiment}
Tan Thongtan and Tanasanee Phienthrakul. 2019.
\newblock \href {https://www.aclweb.org/anthology/P19-2057} {Sentiment
  classification using document embeddings trained with cosine similarity}.
\newblock In \emph{Proceedings of the 57th Annual Meeting of the Association
  for Computational Linguistics: Student Research Workshop}, pages 407--414,
  Florence, Italy. Association for Computational Linguistics.

\bibitem[{Wang et~al.(2019{\natexlab{a}})Wang, Sun, Li, Liu, Si, Zhang, and
  Zhou}]{wang-etal-2019-aspect}
Jingjing Wang, Changlong Sun, Shoushan Li, Xiaozhong Liu, Luo Si, Min Zhang,
  and Guodong Zhou. 2019{\natexlab{a}}.
\newblock \href {https://www.aclweb.org/anthology/P19-1345} {Aspect sentiment
  classification towards question-answering with reinforced bidirectional
  attention network}.
\newblock In \emph{Proceedings of the 57th Annual Meeting of the Association
  for Computational Linguistics}, pages 3548--3557, Florence, Italy.
  Association for Computational Linguistics.

\bibitem[{Wang et~al.(2018)Wang, Mazumder, Liu, Zhou, and Chang}]{WangS:18}
Shuai Wang, Sahisnu Mazumder, Bing Liu, Mianwei Zhou, and Yi~Chang. 2018.
\newblock \href {http://aclweb.org/anthology/P18-1088} {Target-sensitive memory
  networks for aspect sentiment classification}.
\newblock In \emph{Proceedings of the 56th Annual Meeting of the Association
  for Computational Linguistics (Volume 1: Long Papers)}, pages 957--967.
  Association for Computational Linguistics.

\bibitem[{Wang et~al.(2016)Wang, Huang, zhu, and Zhao}]{Wang:16}
Yequan Wang, Minlie Huang, xiaoyan zhu, and Li~Zhao. 2016.
\newblock \href {https://doi.org/10.18653/v1/D16-1058} {Attention-based lstm
  for aspect-level sentiment classification}.
\newblock In \emph{Proceedings of the 2016 Conference on Empirical Methods in
  Natural Language Processing}, pages 606--615. Association for Computational
  Linguistics.

\bibitem[{Wang et~al.(2019{\natexlab{b}})Wang, Sun, Huang, and
  Zhu}]{Wang:2019:ASA:3308558.3313750}
Yequan Wang, Aixin Sun, Minlie Huang, and Xiaoyan Zhu. 2019{\natexlab{b}}.
\newblock \href {https://doi.org/10.1145/3308558.3313750} {Aspect-level
  sentiment analysis using as-capsules}.
\newblock In \emph{The World Wide Web Conference}, WWW '19, pages 2033--2044,
  New York, NY, USA. ACM.

\bibitem[{Weston et~al.(2014)Weston, Chopra, and Bordes}]{Jason:WestonCB14}
Jason Weston, Sumit Chopra, and Antoine Bordes. 2014.
\newblock \href {http://arxiv.org/abs/1410.3916} {Memory networks}.
\newblock \emph{CoRR}, abs/1410.3916.

\bibitem[{Xing et~al.(2019)Xing, Liao, Song, Wang, Zhang, Wang, and
  Huang}]{DBLP:journals/corr/abs-1905-07719}
Bowen Xing, Lejian Liao, Dandan Song, Jingang Wang, Fuzheng Zhang, Zhongyuan
  Wang, and Heyan Huang. 2019.
\newblock \href {http://arxiv.org/abs/1905.07719} {Earlier attention?
  aspect-aware {LSTM} for aspect sentiment analysis}.
\newblock \emph{CoRR}, abs/1905.07719.

\bibitem[{Xu et~al.(2019)Xu, Liu, Shu, and Yu}]{bert2}
Hu~Xu, Bing Liu, Lei Shu, and Philip~S. Yu. 2019.
\newblock \href {http://arxiv.org/abs/1904.02232} {{BERT} post-training for
  review reading comprehension and aspect-based sentiment analysis}.
\newblock \emph{CoRR}, abs/1904.02232.

\bibitem[{Xu and Tan(2018)}]{Xu:18}
Weidi Xu and Ying Tan. 2018.
\newblock \href {http://arxiv.org/abs/1810.10437} {Semi-supervised target-level
  sentiment analysis via variational autoencoder}.
\newblock \emph{CoRR}, abs/1810.10437.

\bibitem[{Xue and Li(2018)}]{weixueGCAE:18}
Wei Xue and Tao Li. 2018.
\newblock \href {http://aclweb.org/anthology/P18-1234} {Aspect based sentiment
  analysis with gated convolutional networks}.
\newblock In \emph{Proceedings of the 56th Annual Meeting of the Association
  for Computational Linguistics (Volume 1: Long Papers)}, pages 2514--2523.
  Association for Computational Linguistics.

\bibitem[{Yang et~al.(2019)Yang, Zhang, Jiang, and Li}]{yang2019aspect}
Chao Yang, Hefeng Zhang, Bin Jiang, and Keqin Li. 2019.
\newblock \href {https://doi.org/10.1016/j.ipm.2018.12.004} {Aspect-based
  sentiment analysis with alternating coattention networks}.
\newblock \emph{Information Processing and Management}, 56:463--478.

\bibitem[{Zhang et~al.(2019)Zhang, Tu, and Zhang}]{zhang-etal-2019-latent}
Liwen Zhang, Kewei Tu, and Yue Zhang. 2019.
\newblock \href {https://www.aclweb.org/anthology/P19-1457} {Latent variable
  sentiment grammar}.
\newblock In \emph{Proceedings of the 57th Annual Meeting of the Association
  for Computational Linguistics}, pages 4642--4651, Florence, Italy.
  Association for Computational Linguistics.

\bibitem[{Zhang and Zhang(2019)}]{zhang-zhang-2019-tree}
Yuan Zhang and Yue Zhang. 2019.
\newblock \href {https://www.aclweb.org/anthology/P19-1342} {Tree communication
  models for sentiment analysis}.
\newblock In \emph{Proceedings of the 57th Annual Meeting of the Association
  for Computational Linguistics}, pages 3518--3527, Florence, Italy.
  Association for Computational Linguistics.

\bibitem[{Zhu and Qian(2018)}]{Zhu:18}
Peisong Zhu and Tieyun Qian. 2018.
\newblock \href {http://aclweb.org/anthology/C18-1092} {Enhanced aspect level
  sentiment classification with auxiliary memory}.
\newblock In \emph{Proceedings of the 27th International Conference on
  Computational Linguistics}, pages 1077--1087. Association for Computational
  Linguistics.

\end{thebibliography}
\bibliographystyle{acl_natbib}

\end{document}